\documentclass[conference,10pt]{IEEEtran}
\usepackage{epsfig,rotating,setspace,latexsym,amsmath,epsf,amssymb,amsfonts,bm,theorem,subfigure,epstopdf}
\usepackage{cite,authblk}
\usepackage{bbm}
\usepackage{color}
\usepackage{mathtools}
\usepackage{algorithm}
\usepackage{algpseudocode}
\usepackage{verbatim}

\algrenewcommand\algorithmicforall{\textbf{foreach}}
\algrenewcommand\algorithmicindent{.8em}

\algnewcommand\algorithmicforeach{\textbf{for each}}
\algdef{S}[FOR]{ForEach}[1]{\algorithmicforeach\ #1\ \algorithmicdo}

\IEEEoverridecommandlockouts
\allowdisplaybreaks

\begin{document}

\title{Personalized Decentralized Multi-Task Learning Over Dynamic Communication Graphs}

\author{Matin Mortaheb \qquad Sennur Ulukus\\
\normalsize Department of Electrical and Computer Engineering\\
\normalsize University of Maryland, College Park, MD 20742\\
\normalsize \emph{mortaheb@umd.edu}  \qquad \emph{ulukus@umd.edu}}

\maketitle

\begin{abstract}
Decentralized and federated learning algorithms face data heterogeneity as one of the biggest challenges, especially when users want to learn a specific task. Even when personalized headers are used concatenated to a shared network (PF-MTL), aggregating all the networks with a decentralized algorithm can result in performance degradation as a result of heterogeneity in the data. Our algorithm uses exchanged gradients to calculate the correlations among tasks automatically, and dynamically adjusts the communication graph to connect mutually beneficial tasks and isolate those that may negatively impact each other. This algorithm improves the learning performance and leads to faster convergence compared to the case where all clients are connected to each other regardless of their correlations. We conduct experiments on a synthetic Gaussian dataset and a large-scale celebrity attributes (CelebA) dataset. The experiment with the synthetic data illustrates that our proposed method is capable of detecting tasks that are positively and negatively correlated. Moreover, the results of the experiments with CelebA demonstrate that the proposed method may produce significantly faster training results than fully-connected networks.
\end{abstract}
 
\section{Introduction}

Decentralized learning (DL) algorithms are able to operate over arbitrary network topologies, in which participants communicate only with their immediate neighbors without a need for communication with a central server. A key challenge in DL is to deal with data heterogeneity: as each agent has its own data, local datasets typically exhibit different distributions. This is especially true for multi-task learning (MTL), where tasks are distributed across different users.

\begin{figure}[t]
 \centerline{\includegraphics[width=1\linewidth]{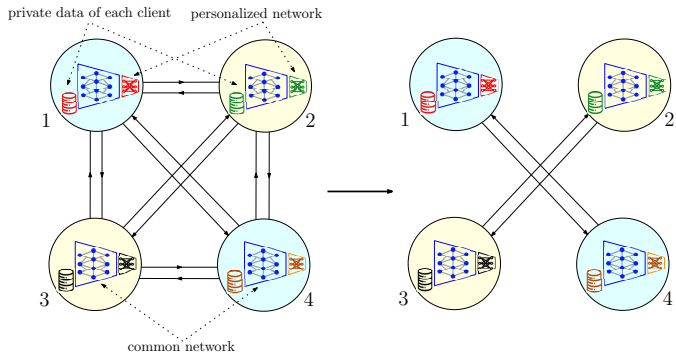}}
  \caption{Dynamic communication graph in personalized decentralized learning framework with a common network (shown in blue) and small personalized headers (shown in red, green, black, orange). Users with the same-color circles have positive correlation, and different colors mean negative correlation.}
  \label{system_model}
  \vspace*{-0.4cm}
\end{figure}

Data heterogeneity is addressed through personalization in federated learning (FL), in which the parameter server and clients train a common base model, and each client additionally trains a small header for its own specific task; shown in Fig.~\ref{system_model} for the DL setting. By using personalization, users can obtain essentially different learning models that are better fitted to their unique data while capturing the common knowledge distilled from other devices' data \cite{arivazhagan2019federated, collins2021exploiting, FedGradNorm}.

Prior to the consideration of heterogeneity as a factor in DL convergence, communication topologies have been entirely characterized by their spectral gaps \cite{wang2019matcha}. The choice of topology, however, has a large impact on heterogeneous settings as observed in \cite{Bellet2021DCliquesCN, neglia2020decentralized, vogels2022beyond}. In the presence of data heterogeneity, choosing a good topology for DL is important. On one hand, clients (tasks) may have adverse effects on each other if the topology of the graph is chosen inappropriately. On the other hand, having a fully connected network can result in a high communication cost. Using a time-varying and data-aware design of the communication network, \cite{le2022refined, dandi2022data} investigate how decentralized SGD performance can be improved in the presence of data heterogeneity. They propose an algorithm to adapt the connectivity matrix (network topology) by minimizing the relative heterogeneity in each round. However, solving an optimization problem that finds the optimal topology in each round results in high computational complexity. 

In order to increase user performance, the topology should be configured to connect similar tasks and to isolate unrelated ones. To identify related tasks, we use a technique called \emph{transference} \cite{fifty2021efficiently}. In MTL \cite{Zhang2017ASO}, transference is a metric to quantify the positive/negative effect of a task's gradient update to the parameters of a shared encoder on another task’s loss during training. In other words, transference metric of $i$ to $j$, $Z_{ij}$, measures the loss of task $j$ before and after applying the gradient update of task $i$ on the shared network. A positive value of $Z_{ij}$ indicates that the update on the shared parameters (by task $i$) results in a smaller loss on task $j$ than the original parameter values. In contrast, a negative value of transference indicates the negative impact of tasks on each other. 

To efficiently design the communication graph, we calculate an approximate transference metric using the gradient updates exchanged among the neighbors. Based on the approximated transference, we dynamically adjust the connectivity matrix $W$ so that the positively correlated tasks are connected together and the negatively correlated tasks are disconnected from each other, thereby, preventing performance degradation. We generate the connectivity matrix by applying the spectral clustering technique \cite{von2007tutorial} to the approximate transference matrix. Our experiments demonstrate that our framework is capable of achieving better performance faster than a fully-connected network in which all users are connected to each other. 

The main contributions of our paper are as follows: 1) We propose a novel algorithm, which dynamically changes the network topology of the users according to their positive/negative correlations. 2) We conduct several experiments on a synthetic Gaussian dataset and a large-scale celebrity attributes (CelebA) dataset \cite{liu2015faceattributes}. By deliberately designing the covariance matrix for the synthetic Gaussian datasets, we show that the framework can detect and cluster correlated tasks correctly. In addition, the experiments with CelebA dataset exhibit faster convergence and improved performance in our framework compared to the fully-connected network.

\section{System Model and Problem Formulation}

We consider a personalized distributed learning (PDL) setting \cite{FedGradNorm, arivazhagan2019federated}, with $N$ clients which are connected via a directed graph. The network of clients aim to solve the following optimization problem associated with an ML problem,
\begin{align} \label{main_optimization}
    \min_{\{ \theta_{s,i} \}_{i=1}^N, \{\theta_{i} \}_{i=1}^N} \left\{ F(\{\theta_{s,i} \}, \{\theta_{i} \}) \triangleq \frac{1}{N} \sum_{i=1}^N \mathcal{L}_i(\mathcal{X}_i^t,\theta_{s,i}^{t},\theta_{i}^{t})\right\}
\end{align}
where $\mathcal{X}_i^t$ is the $i$th user's data at time $t$, and $\mathcal{L}_i(\mathcal{X}_i^t,\theta_{s,i}^{t},\theta_{i}^{t})$ is the task-specific loss obtained by using $\mathcal{X}_i^t$ as an input over the shared network and private head of task $i$. Each client contains a shared network and a private network. The shared network parameters are represented by $\theta_{s,i}^{t}$ and the private head parameters are represented by $\theta_{i}^{t}$, for user $i$ at time $t$. 

The matrix $W$ represents the graph topology among the nodes: $W_{ij} > 0$ means that agent $i$ can communicate with agent $j$, while $W_{ij} = 0$ means that agent $i$ cannot communicate with agent $j$. The matrix $W$ is a mixing matrix, which is composed of values between $0$ and $1$, and is doubly stochastic, i.e., $1^T W=1$ and $W  1=1$, where $1$ is a vector of all ones. Mixing property of the topology matrix $W$ is needed to ensure convergence of an iterative algorithm to solve problem (\ref{main_optimization}). 

Similar to the centralized case, the problem in (\ref{main_optimization}) can be solved through an alternating minimization approach in the decentralized case, using decentralized SGD. First, client $i$ performs $K_1$ local gradient based updates to optimize $\theta_{i}^{t}$, while the global network parameters at client $i$, i.e., $\theta_{s,i}^{t}$, are frozen,  
\begin{align}
    \theta_{i}^{t+1} = \theta_{i}^{t} - \eta \nabla_{\theta_{i}}\mathcal{L}_i(\mathcal{X}_i^t,\theta_{s,i}^{t},\theta_{i}^{t})
\end{align}
Then, by keeping the parameters corresponding to the client-specific head frozen, each node sends the gradient update corresponding to the shared network to its neighbors based on the network topology at time $t$ and aggregates the received updates again based on the network topology specified by the connectivity matrix $W$, $K_2$ times as follows,
\begin{align}
    \theta_{s,i}^{t+1} = \sum_{j=1}^N W_{ij}^t \left(\theta_{s,j}^{t} - \eta \nabla_{\theta_{s,j}}\mathcal{L}_j(\mathcal{X}_j^t,\theta_{s,j}^{t},\theta_{j}^{t})\right)
\end{align}

Now, the objective is to automatically capture the correlations among the users and to dynamically modify the mixing matrix $W$ so that the positively correlated tasks are connected and uncorrelated tasks are disconnected to prevent negative transference. To that end, let us define the quantity $\theta_{s,j|i}^{t+1}$ to represent the updated $j$th shared parameters after a gradient step with respect to loss of task $i$,
\begin{align}\label{condiational_shared_parameter}
    \theta_{s,j|i}^{t+1} = \theta_{s,j}^{t} - \eta \nabla_{\theta_{s,j}}\mathcal{L}_i(\mathcal{X}^t,\theta_{s,j}^{t},\theta_{i}^{t})
\end{align}
where ${X}^t$ is a shared data that each client has for calculating the transference value. 
As a result, the transference of task $i$ on a single task $j$ at time $t$ can be calculated as,
\begin{align}
    Z_{i \rightarrow j}^t = 1- \frac{\mathcal{L}_j(\mathcal{X}^t,\theta_{s,j|i}^{t+1},\theta_{j}^{t+1})}{\mathcal{L}_j(\mathcal{X}^t,\theta_{s,j}^{t},\theta_{j}^{t})}
\end{align}

A positive value for $Z_{i \rightarrow j}^t$ indicates that using task $i$ loss values to update the shared parameters would result in a lower loss for task $j$. Thus, a more positive value for $Z_{i \rightarrow j}^t$ means more correlation among these two tasks. Conversely, having a negative value means that learning task $i$ and $j$ simultaneously would deteriorate the performance of both of the tasks.  

Instead of calculating $\theta_{s,i|i}^{t+1}$ for each user at each time to obtain transference, we can use its first order Taylor series expansion to obtain transference directly from the exchanged gradients among the clients. Let us define the gradient update for the shared network of user $i$ at time $t$ as $g_{s,i}^t = \nabla_{\theta_{s,i}} \mathcal{L}_i(\mathcal{X}^t,\theta_{s,i}^{t},\theta_{i}^{t})$. Then, the first order Taylor series expansion for the transference can be written as,
\begin{align} \label{trans_aprox}
        Z_{i \rightarrow j}^t & = 1- \frac{\mathcal{L}_j(\mathcal{X}^t,\theta_{s,j|i}^{t+1},\theta_{j}^{t+1})}{\mathcal{L}_j(\mathcal{X}^t,\theta_{s,j}^{t},\theta_{j}^{t})} \\
        &\approx 1 - \frac{\mathcal{L}_j(\mathcal{X}^t,\theta_{s,j}^{t},\theta_{j}^{t}) - \langle g_{s,j}^t, g_{s,i}^t \rangle}{\mathcal{L}_j(\mathcal{X}^t,\theta_{s,j}^{t},\theta_{j}^{t})} \\
        & = \frac{\langle g_{s,j}^t, g_{s,i}^t\rangle}{\mathcal{L}_j(\mathcal{X}^t,\theta_{s,j}^{t},\theta_{j}^{t})}
\end{align}

Therefore, instead of calculating $\theta_{s,i|i}^{t+1}$ in each round to obtain $Z_{i \rightarrow j}^t$ value, users can exchange their gradient updates obtained via the shared data ${X}^t$ to more efficiently compute $Z_{i \rightarrow j}^t$. After calculating $Z^t \in \mathbb{R}^{N \times N}$ matrix, we must convert it to a doubly stochastic mixing matrix $W^t$ to ensure the convergence of the algorithm. We convert $Z^t$ to a doubly stochastic matrix by clustering the $Z^t$ matrix. 

We use the \emph{spectral clustering} technique \cite{von2007tutorial} to perform clustering over a communication graph expressed by matrix $Z^t$. In spectral clustering, we use the eigenvalues of the graph Laplacian to find the appropriate clusters. To calculate the graph Laplacian, let us define matrix $D$ as the degree matrix, which is a diagonal matrix where the $(i,i)$th entry indicates the degree of node $i$ (the number of edges connected to node $i$). Then, the graph Laplacian can be calculated as $L = D - Z$. In the graph Laplacian matrix, diagonal entries are the degrees of the nodes, and off-diagonal entries are the negative edge weights. Finally, we calculate the eigenvalues of the graph Laplacian matrix $L$. 

By sorting the eigenvalues, we see that the number of 0 eigenvalues corresponds to the number of connected components in the graph. Also, an eigenvalue with a small value, e.g., $\text{eig}(L) \leq 1$, indicates that there is almost a separation of the two components. Therefore, we can determine the number of clusters by calculating the number of graph Laplacian eigenvalues which have values less than 1. The vectors associated with those eigenvalues contain information as to how to segment the network. Finally, we perform $k$-means on those vectors in order to obtain the labels for the nodes. The spectral clustering algorithm is given in Algorithm~\ref{alg:spectral_clustering}.
\begin{algorithm}[h]
    \caption{Unnormalized spectral clustering \cite{von2007tutorial}}
    \label{alg:spectral_clustering}
\begin{algorithmic}[1]
\State {\bfseries Input:} Similarity matrix $Z$, number $k$ of clusters to construct.
\State Compute the unnormalized Laplacian $L$.
\State Compute the first $k$ eigenvectors $u_1, \ldots, u_k$ of $L$.
\State Let $U\in \mathcal{R}^{n \times k}$ be the matrix containing the vectors $u_1, \ldots, u_k$ as columns.
\State For $i = 1, \ldots, n$, let $y_i \in \mathbb{R}^{k}$ be the vector corresponding to the $i$th row of $U$.
\State Cluster the points $\{y_i\}_{i = 1, \ldots, n}$ in $\mathbb{R}^{k}$ with the $k$-means algorithm into clusters $C_1, \ldots, C_k$. 
\State {\bfseries Output:} Clusters $A_1, \ldots, A_k$ with $A_i = \{j| y_j \in C_i\}$.
\end{algorithmic}
\end{algorithm}

Let $D_l$ be the number of nodes in cluster $A_l$ for $l \in 1, \ldots, k$. Then, $W^t$ can be calculated as,
\begin{align} \label{spectral}
    W^t(i,j)= 
    \begin{cases}
        \frac{1}{d_l},& \text{if } i,j \in A_l\\
        0,              & \text{otherwise}
\end{cases}
\end{align}
This method ensures that $W^t$ remains doubly-stochastic. To prevent error due to the tolerance of calculating transference, the $Z^t$ matrix is averaged over each $H$ epochs and then $W^t$ is calculated by using spectral clustering.

Finally, in each training step, we perform a decentralized SGD step to exchange and update the shared network across users connected via an edge in the derived topology at time $t$, $W^t$. The overall algorithm is summarized in Algorithm~\ref{alg:first_method}. 

\begin{algorithm}[h]
    \caption{Training with our proposed algorithm}
    \label{alg:first_method}
\begin{algorithmic}[1]
\State {\bfseries Input:} step sizes $\eta$, initialization $\theta_{s}$, $\{\theta_{i}|i\in N\}$, $H$, $K_2$.
\For{t = 1, \ldots, $T$}
\ForEach {$i \in N $ (in parallel)}
\State Compute task-loss $\mathcal{L}_i(\mathcal{X}_i^t,\theta_{s,i}^{t},\theta_{i}^{t})$
\State $\theta_{i}^{t+1} \leftarrow \theta_{i}^{t} - \eta \nabla_{\theta_i}\mathcal{L}_i(\mathcal{X}_i^t,\theta_{s,i}^{t},\theta_{i}^{t})$
\EndFor
\ForEach{$i,j \in N $ (in parallel)}
%\State $\theta_{s,i|i}^{t+1} = \theta_{s,i}^{t} - \eta \nabla_{\theta_{s,i}}\mathcal{L}_i(\mathcal{X}^t,\theta_{s,i}^{t},\theta_{i}^{t})$
\State Calculating $Z_{i \rightarrow j}^t$ using (\ref{trans_aprox})
\EndFor
\State Calculating mixing matrix $V^{t}$ from $Z^t$ using (\ref{spectral})
\State $W_{temp}^{t} = W_{temp}^{t} + V^{t}$
\If{$t = H$} 
\State $W^t = W_{temp}^t/H$
\State $W_{temp} = 0$
\EndIf
\For{$k$= 1, \ldots, $K_2$}
\ForEach {$i \in N$ (in parallel)}
\State $\theta_{s,i}^{t+\frac{1}{2}} \leftarrow \theta_{s,i}^{t} - \eta \nabla_{\theta_s}\mathcal{L}_i(\mathcal{X}_i^t,\theta_{s,i}^{t},\theta_{i}^{t})$
\State $\theta_{s,i}^{t+1} \leftarrow \sum_{j=1}^N W_{ij}^t \theta_{s,j}^{t+\frac{1}{2}}$
\EndFor
\EndFor
\EndFor
\end{algorithmic}
\end{algorithm}

\section{Convergence Analysis}

Our convergence analysis follows \cite{koloskova2019decentralized,koloskova2020unified}. We assume that each worker's objective function $f_i: \mathbb{R}^{d+d_i} \rightarrow \mathbb{R}$ for all $i$ is $L$-smooth and $\mu$-strongly convex and that the variance on each worker is bounded. We also assume that the connectivity matrix (mixing matrix) is doubly-stochastic. Therefore, the convergence analysis follows from \cite[Thm.~4]{koloskova2019decentralized}. Further, \cite{koloskova2020unified} uses much weaker assumptions to prove convergence.

\section{Experimental Results}
We compare the task losses achieved by naive fully connected network and our proposed dynamic communication graph algorithm.

\subsection{Dataset Specifications}
We use the following two datasets for our experiments:

\emph{Synthetic Gaussian vector dataset} that contains 30,000 training data, 10,000 test data, and 10,000 shared data. Each data point in this dataset is a Gaussian vector of size 10, i.e., $x_i \in \mathbb{R}^{10}$, with mean of $\mu \in \mathbb{R}^{10}$ and covariance of $\Sigma \in \mathbb{R}^{10 \times 10}$. Same as Fig.~\ref{system_model}, the covariance matrix is purposefully designed in such a way that the first and fourth coordinates are positively correlated, and they are negatively correlated with the second and third coordinates. Conversely, the second and third coordinates have a positive correlation, while they both have negative correlations with the first and forth tasks. We consider four attributes for each sample in our experiments. The first four coordinates with purposefully assigned correlations are selected as those four attributes. For each sample, the attributes are 1 if the corresponding coordinate is greater than the assigned mean and 0 otherwise. This synthetic dataset is designed to test the effectiveness of our proposed algorithm to connect the tasks who have positive correlation among themselves and disconnect those who have negative correlations. The training data points are distributed among clients, but each client has access only to one of the attributes (coordinates). 

\begin{figure}[t]
\centerline{\includegraphics[width=0.8\linewidth]{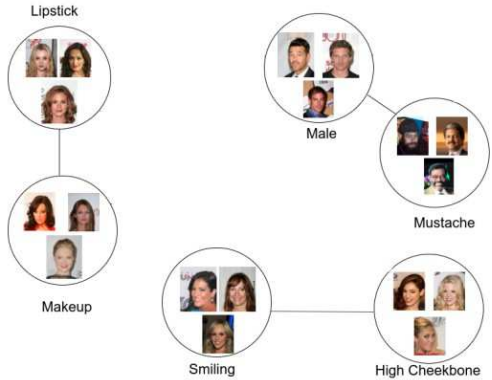}}
  \caption{An expected communication graph (correlation) among the chosen attributes from CelebA dataset.}
\centering
\label{celebA_graph}
\end{figure}

\emph{CelebA dataset} \cite{liu2015faceattributes} that contains 200,000 images, where each image contains 40 attributes. In our experiment, we only use 6 of the attributes, namely, male, mustache, high-cheekbones, smiling, heavy-makeup, and wearing lipsticks. Therefore, the dataset is divided into 6 parts, and each user has access to a single attribute of the given images. The chosen attributes and the expected correlation among attributes is shown in Fig.~\ref{celebA_graph}.

\subsection{Hyperparameters and Model Specifications}

We use Adam optimizer with learning rate $\eta = 2 \times 10^{-5}$ to train the shared and personalized networks for both datasets. In the synthetic Gaussian dataset, we use a shared encoder as explained in Table~\ref{share_model_table}, and each client also has a simple linear layer as a personalized network that maps the shared network’s output to the corresponding prediction value. We use cross-entropy as a loss function for the classification task. 

\begin{table}[h!]
\begin{center}
\begin{sc}
\begin{tabular}{|c|}
\hline
Shared Network\\
\hline
FC(10, 64)\\
FC(64, 128)\\
FC(128, 256)\\
FC(256, 512)\\
FC(512, 256)\\
FC(256, 128)\\
\hline
\end{tabular}
\end{sc}
\end{center}
\caption{Shared network model.}
\label{share_model_table}
\vspace*{-0.4cm}
\end{table}

In CelebA dataset, we use resnet-18 as a shared network for all the users, and a simple 2-layer network for each user to map the output of the shared network to the corresponding classification task. In both experiments, we perform $K_2 = 2$ decentralized-SGD protocols in each epoch. Also, we average the transference metric through 5 epochs before modifying the connectivity matrix, i.e., $H = 5$.

\subsection{Results and Analysis}

\begin{figure}[]
 	\begin{center}
 	\subfigure[]{%
 	\includegraphics[width=0.49\linewidth]{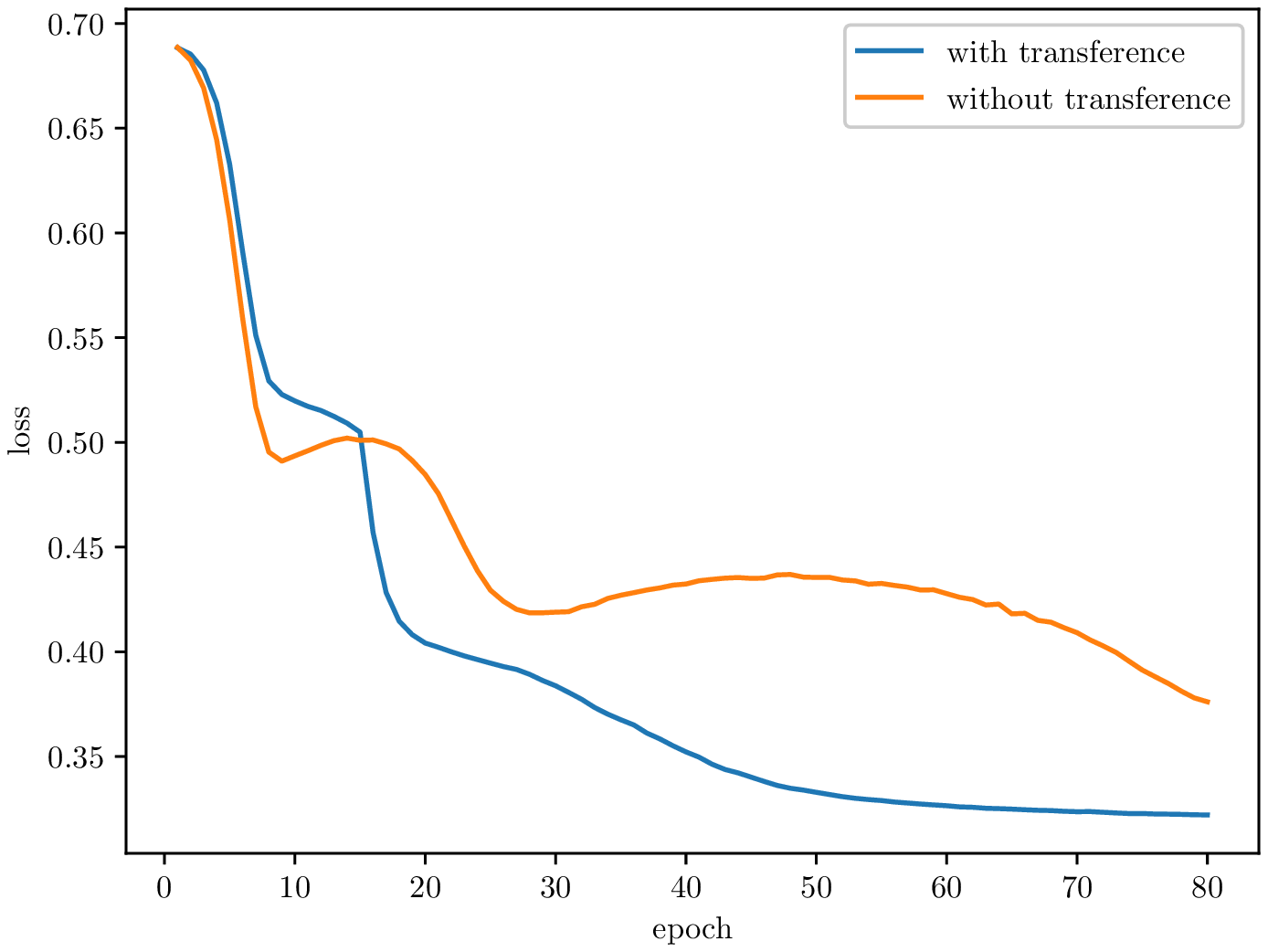}}
 	\subfigure[]{%
 	\includegraphics[width=0.49\linewidth]{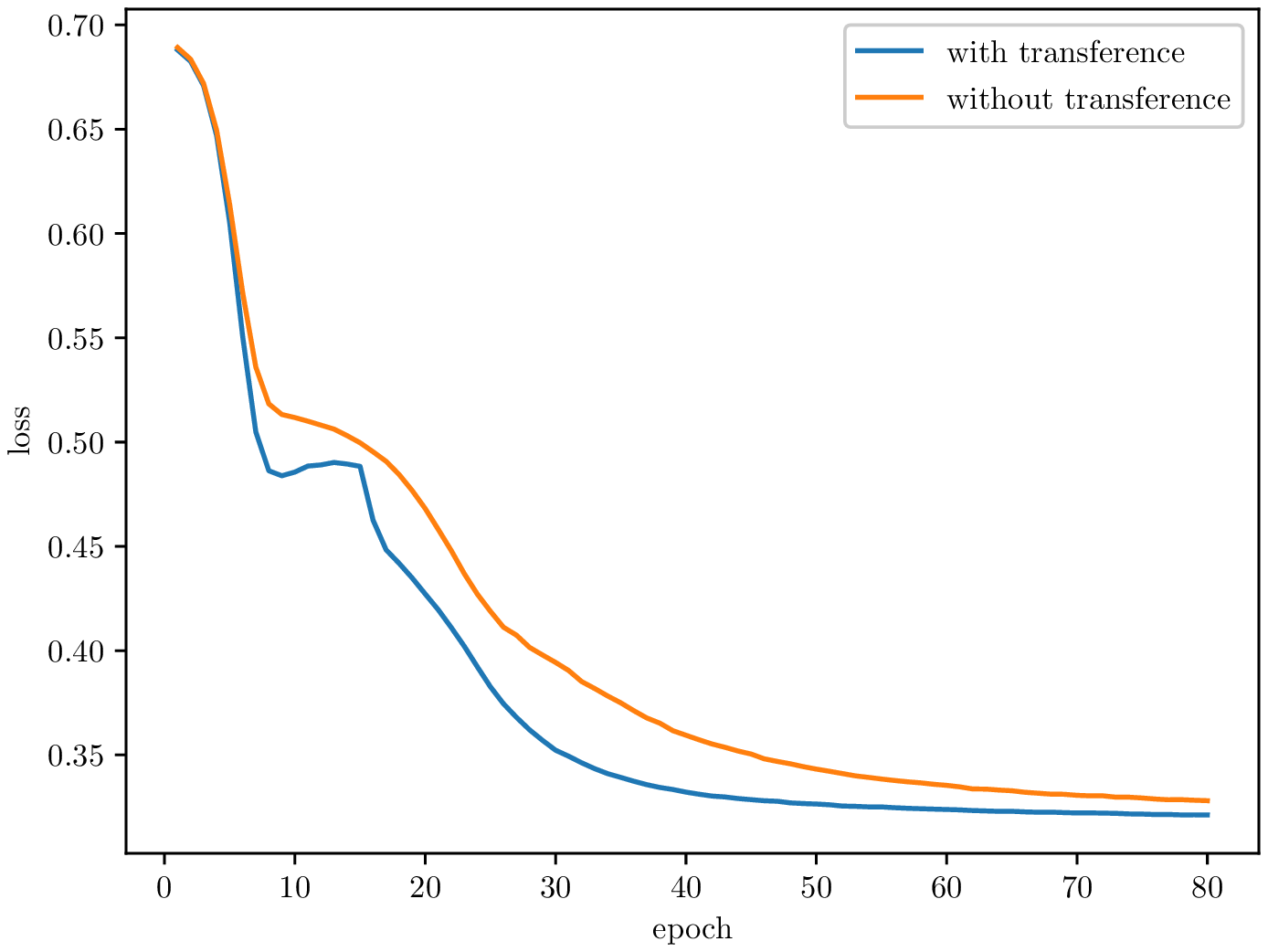}}\\ 
 	\subfigure[]{%
 	\includegraphics[width=0.49\linewidth]{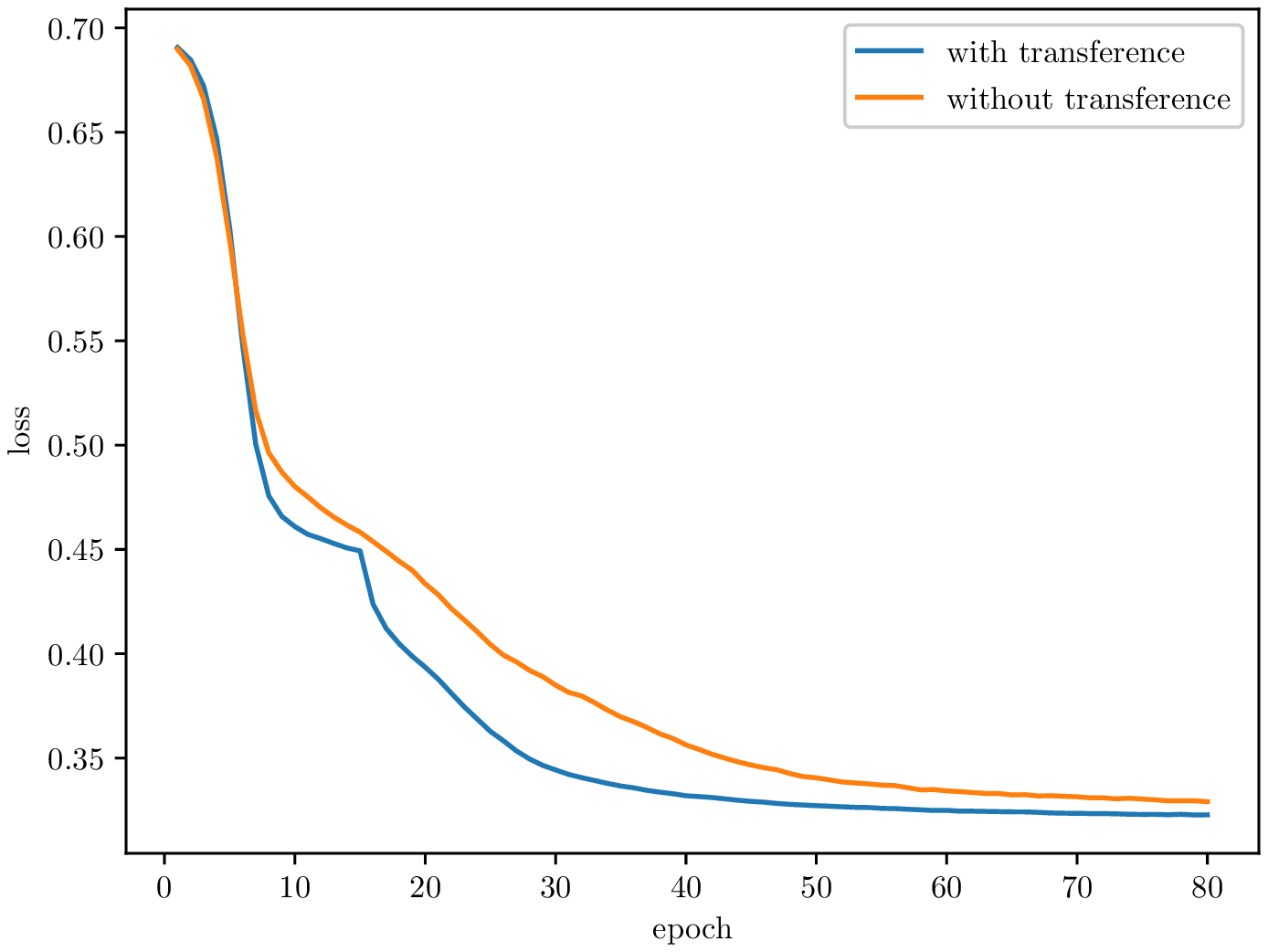}}
 	\subfigure[]{%
 	\includegraphics[width=0.49\linewidth]{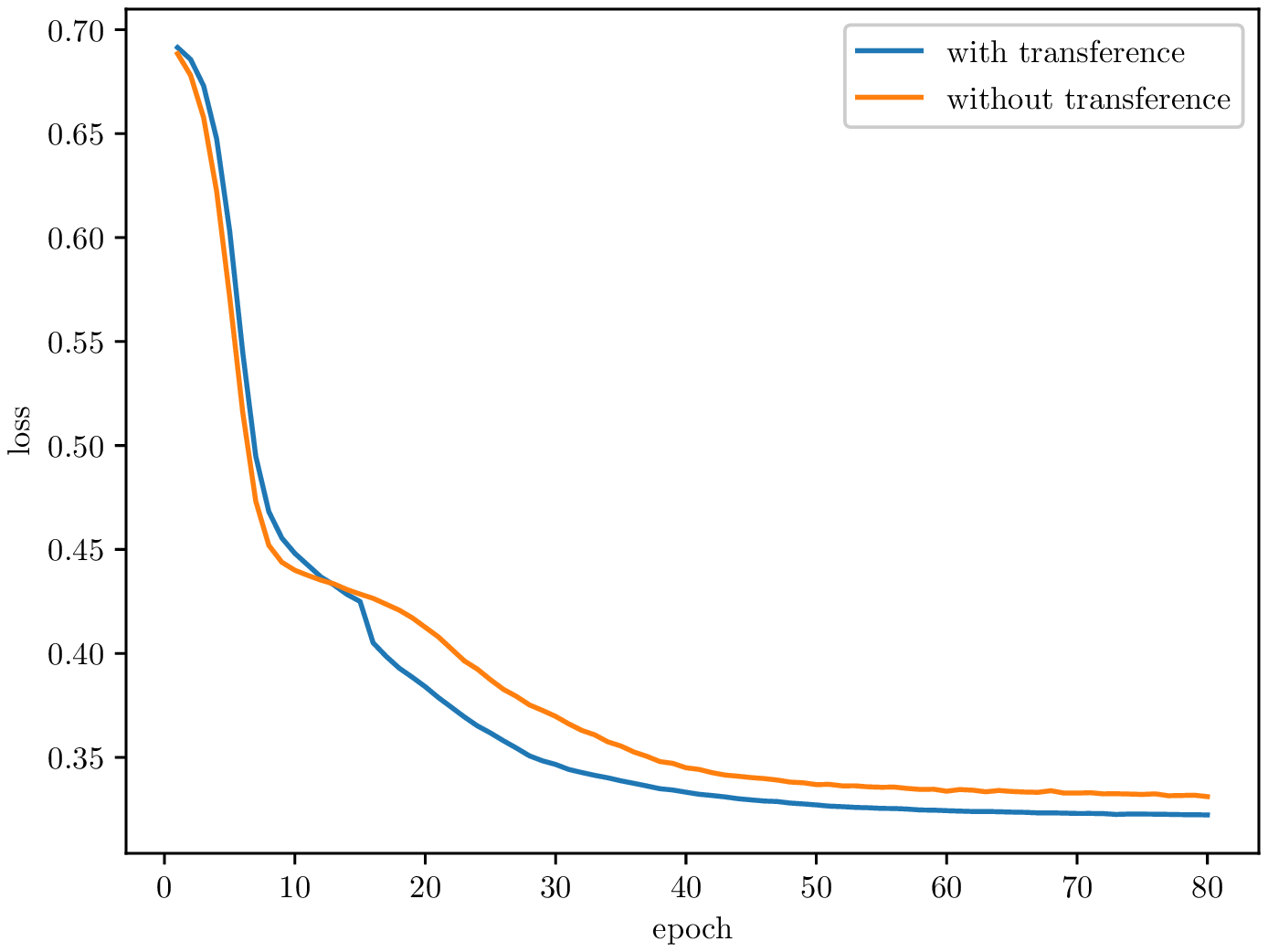}}
 	\end{center}
 	\caption{Comparison of task losses achieved via dynamic communication graph using transference and naive fully-connected case for synthetic dataset.}
 	\label{loss_synthetic}
\end{figure}

We begin with the synthetic dataset. As shown in Fig.~\ref{loss_synthetic}, by capturing the true correlation among users with the transference method at epoch 15, the connectivity matrix clusters users properly based on their correlation, which results in improved performance as compared to the case where all users are connected regardless of their correlation to the other tasks. For tasks~2-4, our proposed method converges approximately at epoch 30, while the naive fully-connected method converges at epoch 60. A faster convergence is more evident in task~1 where our method converges after 45 epochs, while a naive fully-connected network does not converge satisfactorily even after 80 epochs.

\begin{figure}[]
 	\begin{center}
 	\subfigure[]{%
 	\includegraphics[width=0.49\linewidth]{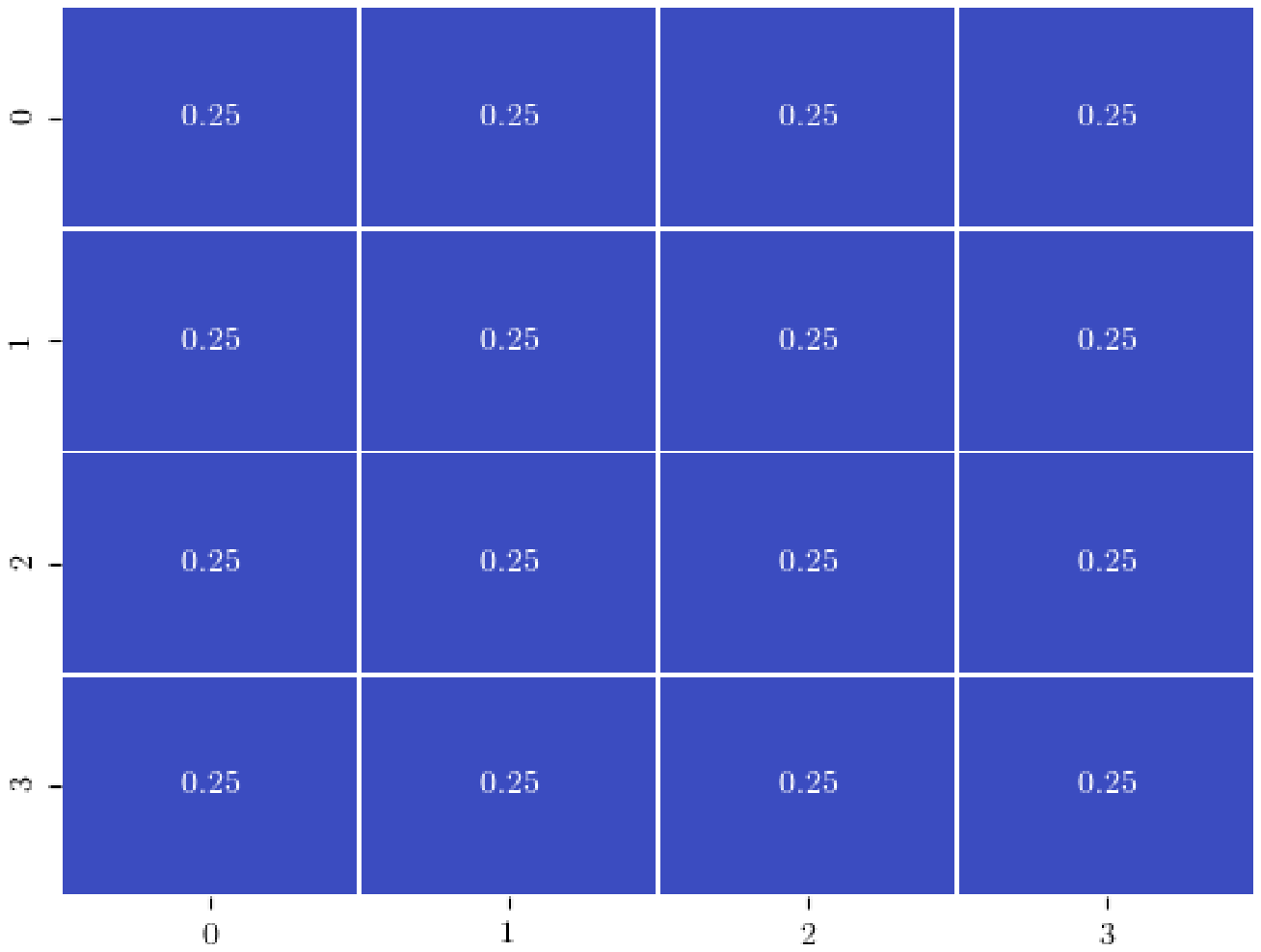}}
 	\subfigure[]{%
 	\includegraphics[width=0.49\linewidth]{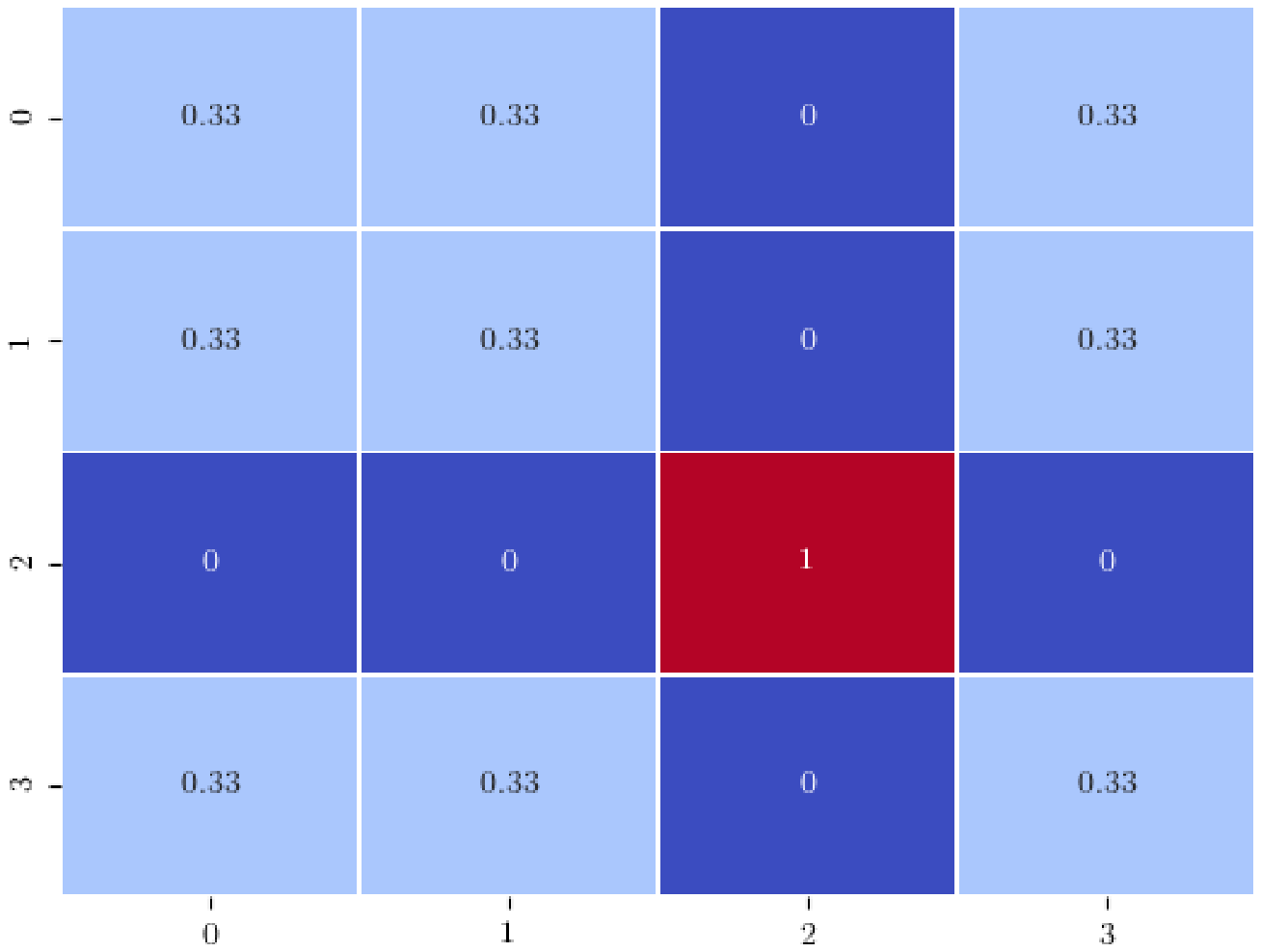}}\\ 
 	\subfigure[]{%
 	\includegraphics[width=0.49\linewidth]{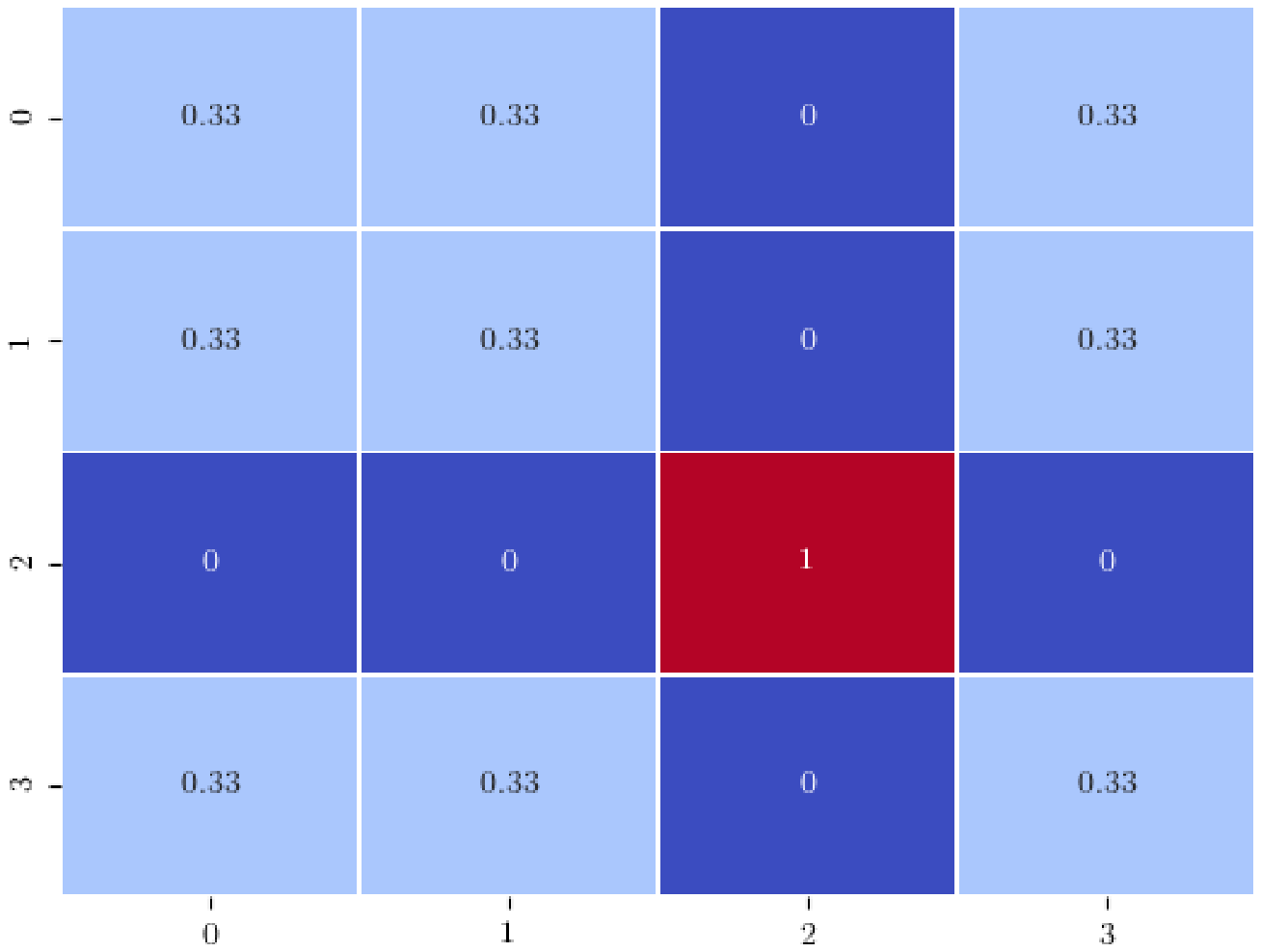}}
 	\subfigure[]{%
 	\includegraphics[width=0.49\linewidth]{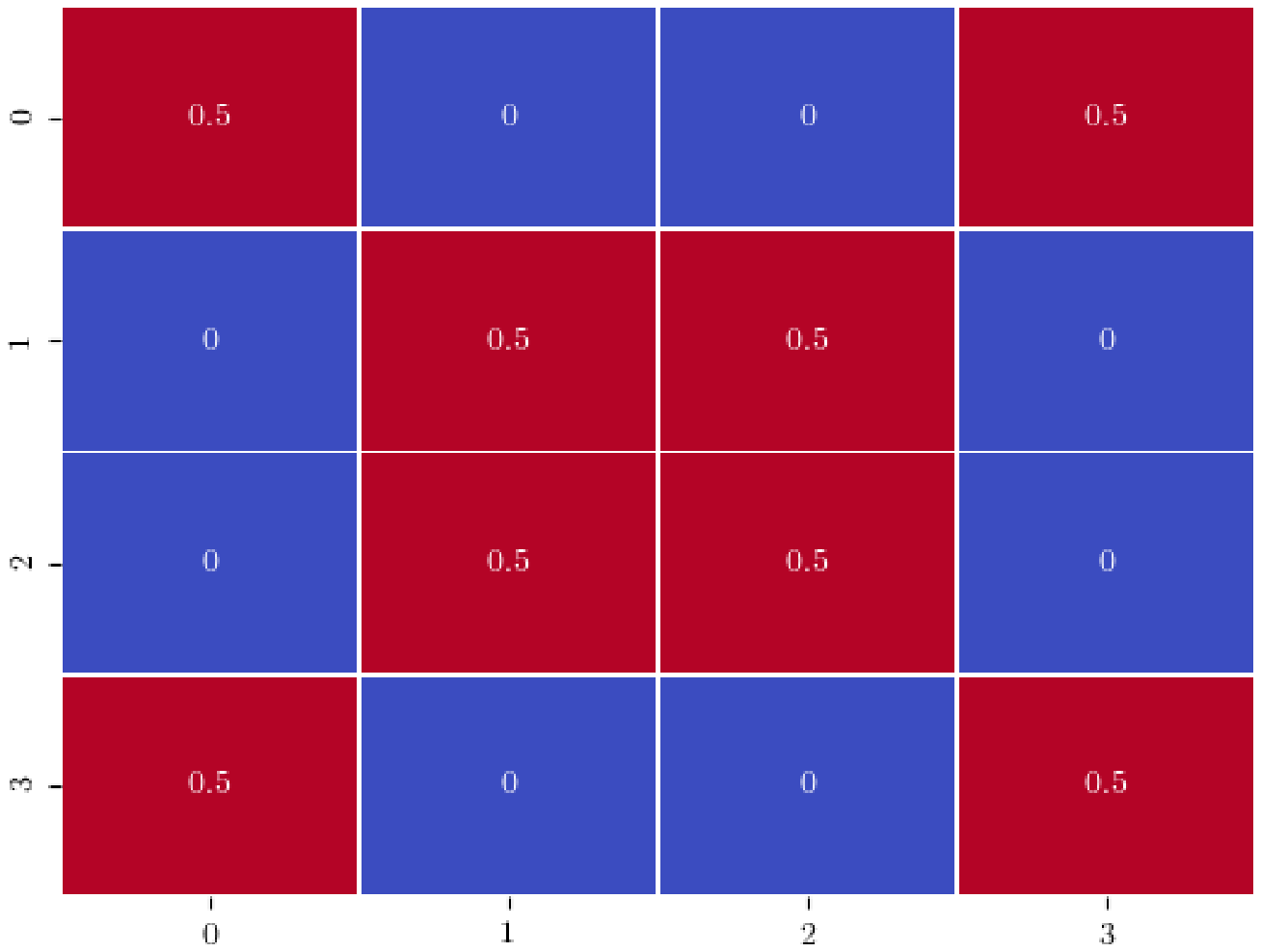}}
 	\end{center}
 	\caption{Heatmap of the connectivity matrix achieved by the transference method over different epochs, (a) epoch 1, (b) epoch 6, (c) epoch 11, (d) epoch 16.}
 	\label{heatmap_synthetic}
\end{figure}

The heatmap of the dynamic topology is shown in Fig.~\ref{heatmap_synthetic}. According to our proposed method, the connectivity matrix initially starts with a fully-connected network, and the topology changes every five epochs based on the calculated transference. In epoch~15, the topology finally captures the correct transference as purposefully designed as in Fig.~\ref{system_model}. Ultimately, the weights are uniformly distributed among tasks~1, 4 and tasks~2, 3 that have positive correlations between them.

\begin{figure}[]
 	\begin{center}
 	\subfigure[]{%
 	\includegraphics[width=0.49\linewidth]{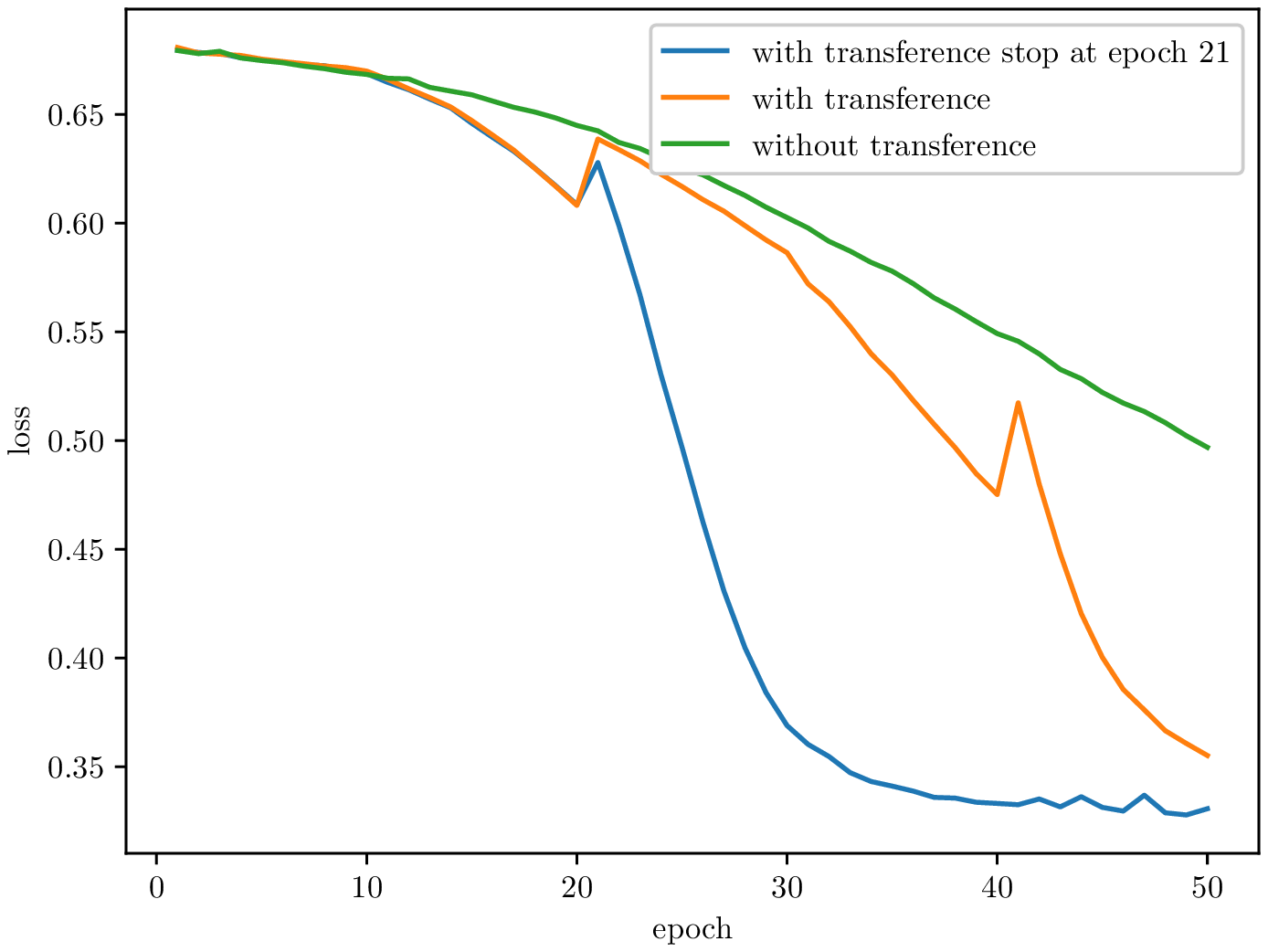}}
 	\subfigure[]{%
 	\includegraphics[width=0.49\linewidth]{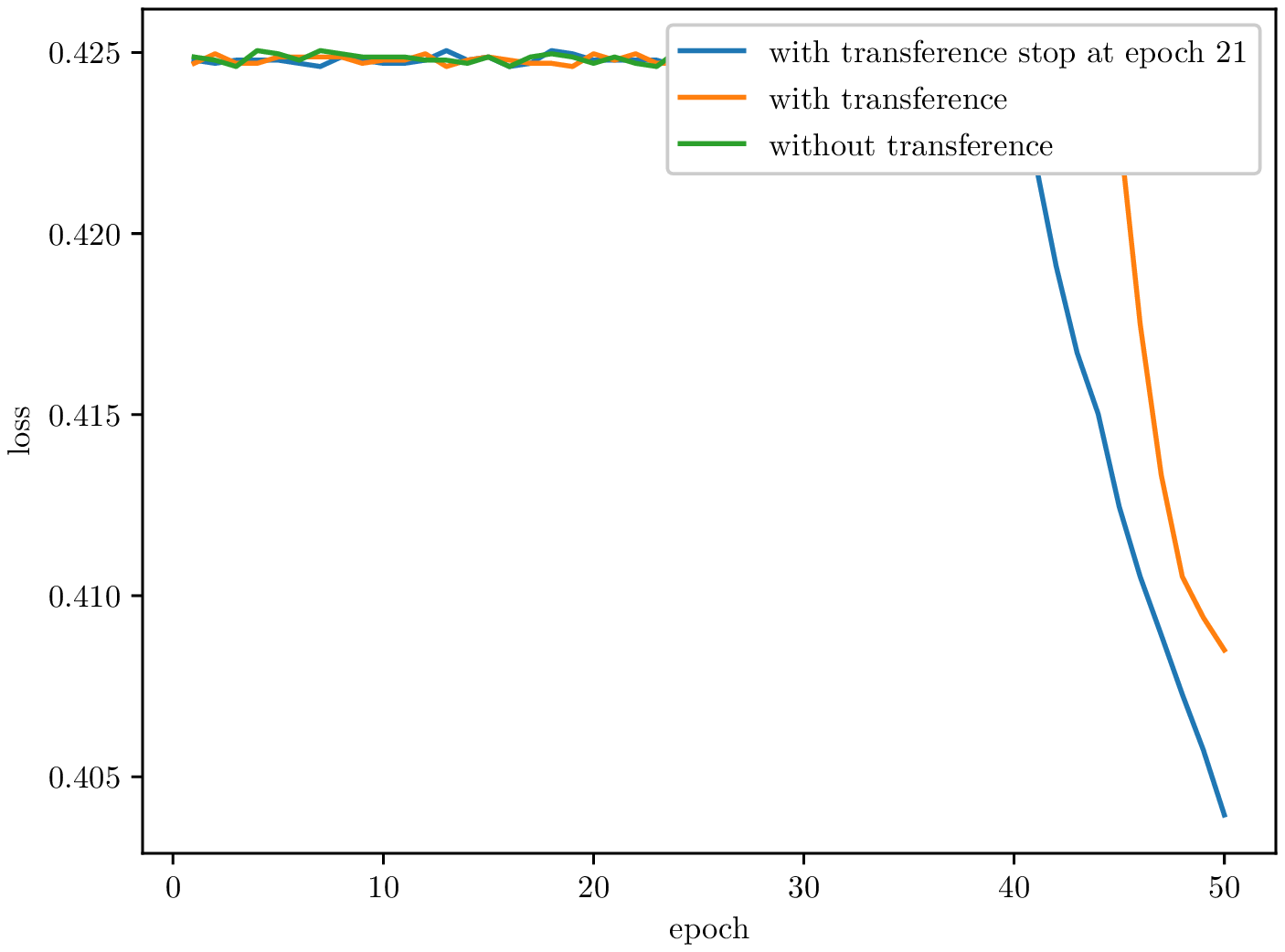}}\\ 
 	\subfigure[]{%
 	\includegraphics[width=0.49\linewidth]{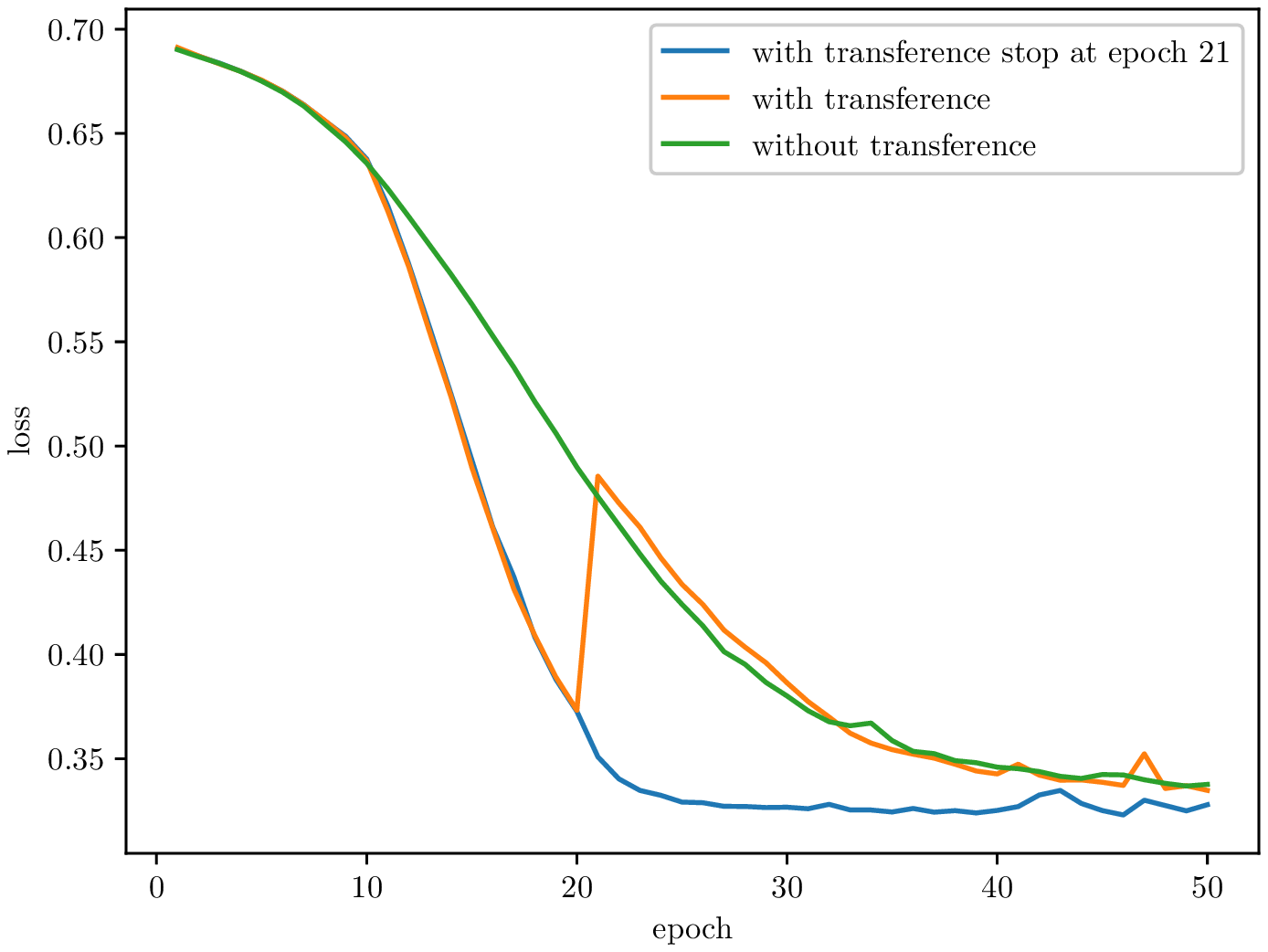}}
 	\subfigure[]{%
 	\includegraphics[width=0.49\linewidth]{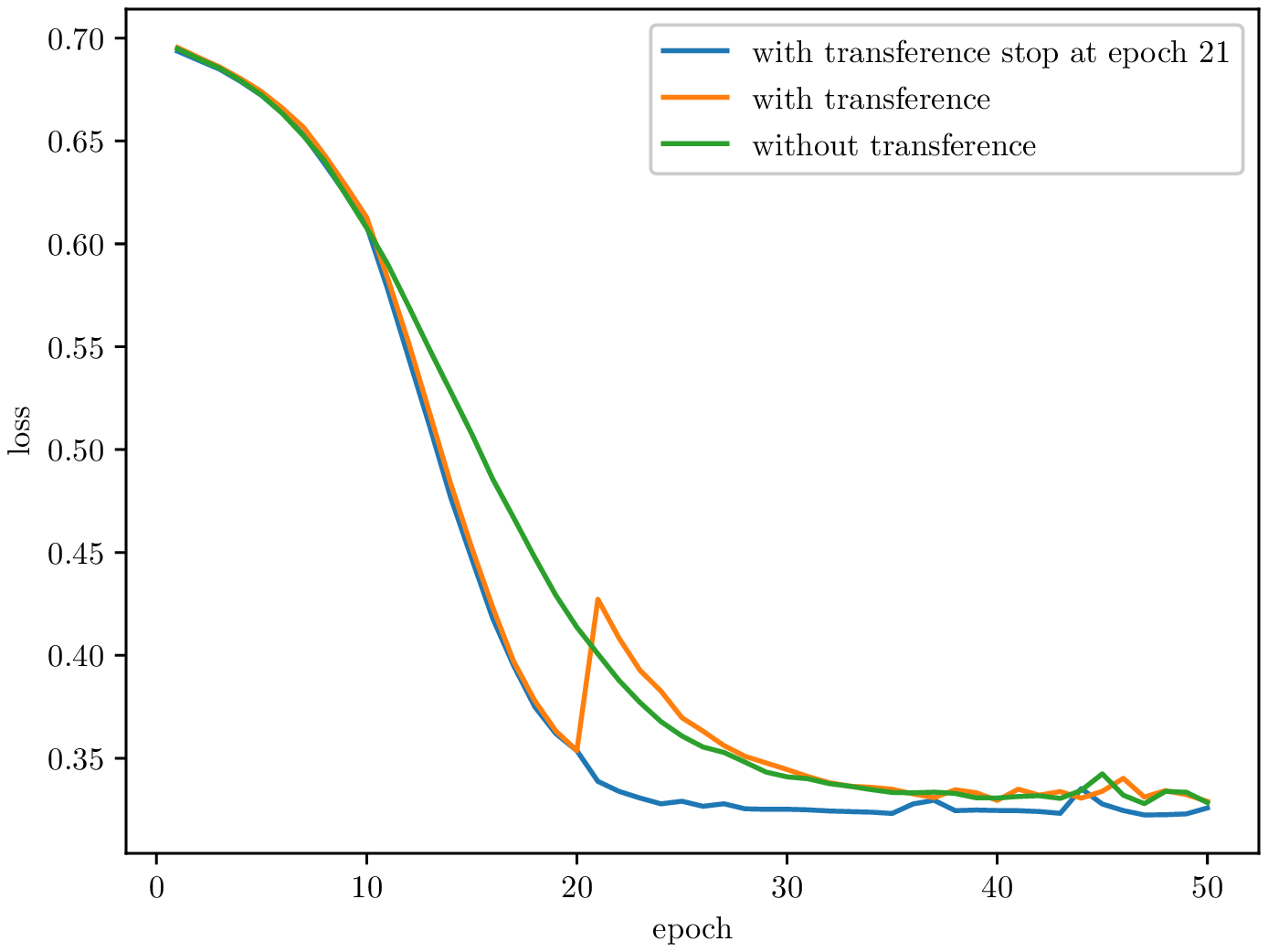}}\\
 	\subfigure[]{%
 	\includegraphics[width=0.49\linewidth]{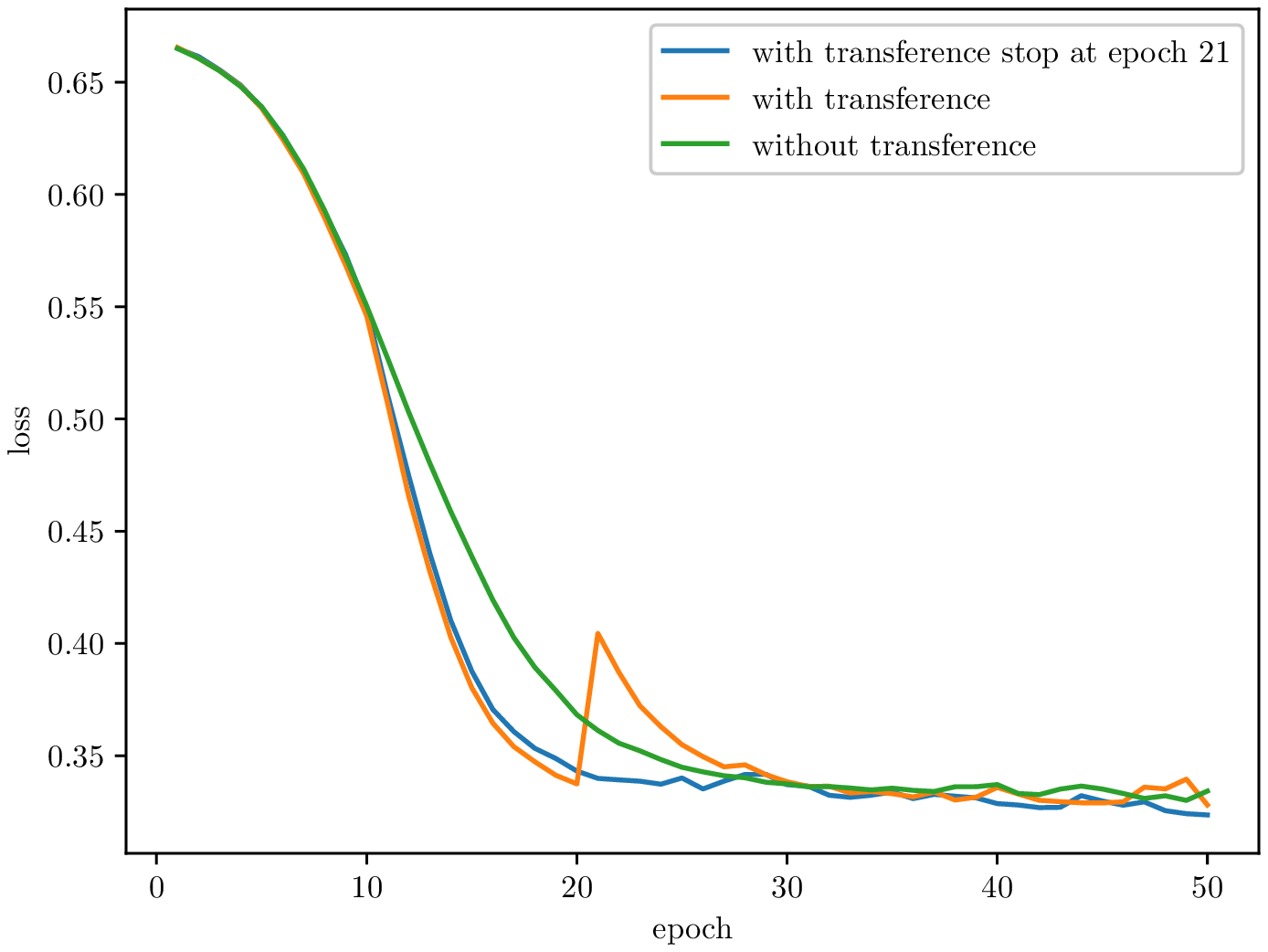}}
 	\subfigure[]{%
 	\includegraphics[width=0.49\linewidth]{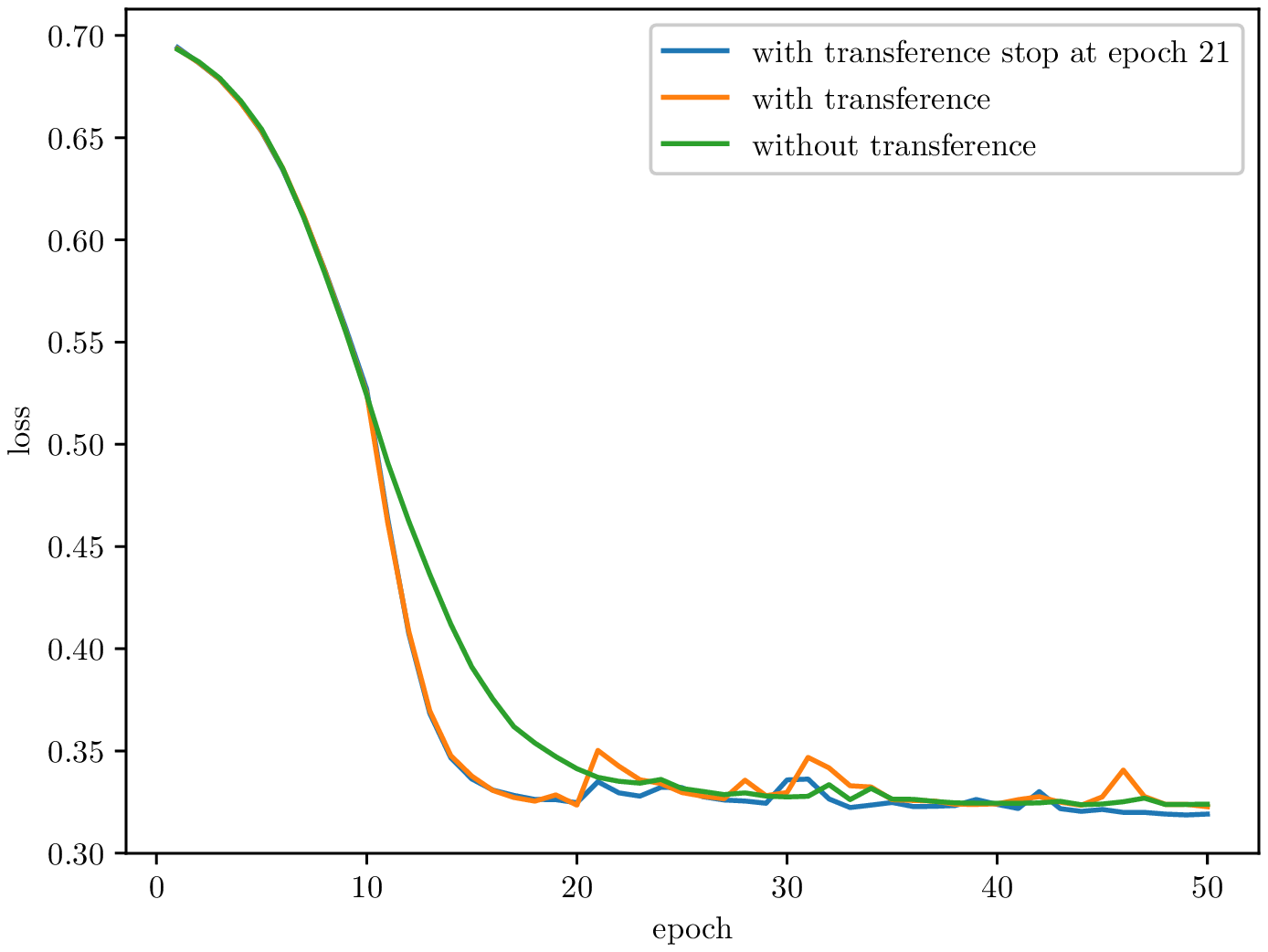}}
 	\end{center}
 	\caption{Comparison of task losses achieved via dynamic communication graph using transference and naive fully-connected case for celebA dataset.}
 	\label{loss_celeba}
\end{figure}

Next we examine the proposed method on a larger and more realistic dataset, CelebA. As shown in Fig.~\ref{loss_celeba}, the results demonstrate the superiority of our method compared to a naive fully-connected network. As part of this experiment, we stopped changing the connectivity matrix at epoch~21 to observe the effect of changing the topology on the initial training phase compared with the whole training process. Our results indicate that the training speed increases when the topology is not modified after sufficiently capturing the correlation of tasks during the initial training phase. Therefore, changing the topology until the end of the training may result in drifting from the optimal performance and slower convergence. However, in the majority of our experiments, our proposed method converged faster in both early stopping and permanent topology change cases.

As shown in Fig.~\ref{loss_celeba}, as a result of our proposed method, the loss of task~1 drops with a higher slope after epoch~20, while further topology change can reduce the slope of dropping losses due to the sudden changes that may occur in the topology of the graph. Hence, our method with early stopping topology change converges at approximately epoch~40, while the transference method with a permanent topology change converges at epoch~50, and the naive fully-connected network does not converge satisfactorily even at epoch~50. The second task (learning mustache) converges quickly even from the very first epochs since it is a simple task to learn. Nevertheless, after around epoch~40, when the other correlated task converges appropriately, the loss decreases due to the changed communication topology. As a result of our proposed method, we have observed that task~3-6 have converged faster at around epoch~10, while the losses for the fully-connected tasks have started to decline much more slowly. The permanent modification of the topology matrix, however, may result in sudden divergence from the convergence and may delay the pace of convergence. 

\bibliographystyle{unsrt}
\bibliography{reference}
\end{document}